\documentclass[sigconf]{acmart}

\usepackage{enumitem}
\usepackage{bm}
\usepackage{multirow}
\usepackage{multicol}
\usepackage{amsfonts}
\usepackage{graphicx}
\usepackage{tabularx}
\usepackage{subfigure}
\usepackage{colortbl}
\usepackage[capitalise]{cleveref}
\Crefname{figure}{Fig.}{Figs.}
\Crefname{table}{Tab.}{Tabs.}
\Crefname{equation}{Eq.}{Eqs.}
\usepackage[most]{tcolorbox}
\usepackage[framemethod=tikz]{mdframed}
\mdfdefinestyle{customstyle}{
  linecolor=gray!100,
  linewidth=3pt,
  innerleftmargin=3pt,
  topline=false,
  rightline=false,
  bottomline=false,
  leftline=true,
  innerrightmargin=3pt,
  innertopmargin=3pt,
  innerbottommargin=3pt,
  backgroundcolor=gray!8
}
\newenvironment{custommdframed}
  {\begin{mdframed}[style=customstyle]}
  {\end{mdframed}}
\setlist[itemize,enumerate]{leftmargin=*, topsep=0pt}

\newcommand{\modelname}{\texttt{ColaCare}}

\AtBeginDocument{
  \providecommand\BibTeX{{
    \normalfont B\kern-0.5em{\scshape i\kern-0.25em b}\kern-0.8em\TeX}}}

\copyrightyear{2025}
\acmYear{2025}
\setcopyright{acmlicensed}
\acmConference[WWW '25] {Proceedings of the ACM Web Conference 2025}{April 28--May 2, 2025}{Sydney, NSW, Australia.}
\acmBooktitle{Proceedings of the ACM Web Conference 2025 (WWW '25), April 28--May 2, 2025, Sydney, NSW, Australia}
\acmISBN{979-8-4007-1274-6/25/04}
\acmDOI{10.1145/3696410.3714877}

\begin{document}

\title[\modelname{}: Enhancing EHR Modeling through LLM-Driven Multi-Agent Collaboration]{\modelname{}: Enhancing Electronic Health Record Modeling through Large Language Model-Driven Multi-Agent Collaboration}

\author{Zixiang Wang}
\orcid{0009-0000-1257-9580}
\authornote{Equal contribution.}
\affiliation{
  \institution{Peking University}
  \city{Beijing}
  \country{China}}
\email{wangzx@stu.pku.edu.cn}

\author{Yinghao Zhu}
\orcid{0000-0002-2640-6477}
\authornotemark[1]
\affiliation{
  \institution{Peking University}
  \city{Beijing}
  \country{China}}
\email{yhzhu99@gmail.com}

\author{Huiya Zhao}
\orcid{0009-0009-8660-801X}
\authornotemark[1]
\affiliation{
  \institution{Peking University}
  \city{Beijing}
  \country{China}}
\email{zhaohuiya@stu.pku.edu.cn}

\author{Xiaochen Zheng}
\orcid{0009-0007-9714-2193}
\affiliation{
  \institution{ETH Zurich}
  \city{Zurich}
  \country{Switzerland}}
\email{xzheng@ethz.ch}

\author{Dehao Sui}
\orcid{0009-0000-9081-059X}
\affiliation{
  \institution{Peking University}
  \city{Beijing}
  \country{China}}
\email{dehaosui1@gmail.com}

\author{Tianlong Wang}
\orcid{0009-0002-7292-6868}
\affiliation{
  \institution{Peking University}
  \city{Beijing}
  \country{China}}
\email{tianlong.wang@stu.pku.edu.cn}

\author{Wen Tang}
\orcid{0000-0002-2263-2979}
\affiliation{
  \institution{Peking University Third Hospital}
  \city{Beijing}
  \country{China}}
\email{tanggwen@126.com}

\author{Yasha Wang}
\orcid{0000-0002-8026-9688}
\authornotemark[2]
\affiliation{
  \institution{Peking University}
  \city{Beijing}
  \country{China}}
\email{wangyasha@pku.edu.cn}

\author{Ewen Harrison}
\orcid{0000-0002-5018-3066}
\affiliation{
  \institution{University of Edinburgh}
  \city{Edinburgh, Scotland}
  \country{UK}}
\email{ewen.harrison@ed.ac.uk}

\author{Chengwei Pan}
\orcid{0000-0003-0497-7903}
\affiliation{
  \institution{Beihang University}
  \city{Beijing}
  \country{China}
}
\email{pancw@buaa.edu.cn}

\author{Junyi Gao}
\orcid{0000-0002-4951-8682}
\authornotemark[2]
\affiliation{
  \institution{University of Edinburgh}
  \city{Edinburgh}
  \state{Scotland}
  \country{UK}}
\affiliation{
  \institution{Health Data Research UK}
  \city{London}
  \country{UK}}
\email{junyi.gao@ed.ac.uk}

\author{Liantao Ma}
\orcid{0000-0001-5233-0624}
\authornote{Corresponding author.}
\affiliation{
  \institution{Peking University}
  \city{Beijing}
  \country{China}}
\email{malt@pku.edu.cn}

\renewcommand{\shortauthors}{Zixiang Wang et al.}


\begin{abstract}
We introduce \modelname{}, a framework that enhances Electronic Health Record (EHR) modeling through multi-agent collaboration driven by Large Language Models (LLMs). Our approach seamlessly integrates domain-specific expert models with LLMs to bridge the gap between structured EHR data and text-based reasoning. Inspired by the Multidisciplinary Team (MDT) approach used in clinical settings, \modelname{} employs two types of agents: \texttt{DoctorAgent}s and a \texttt{MetaAgent}, which collaboratively analyze patient data. Expert models process and generate predictions from numerical EHR data, while LLM agents produce reasoning references and decision-making reports within the MDT-driven collaborative consultation framework. The \texttt{MetaAgent} orchestrates the discussion, facilitating consultations and evidence-based debates among \texttt{DoctorAgent}s, simulating diverse expertise in clinical decision-making. We additionally incorporate the Merck Manual of Diagnosis and Therapy (MSD) medical guideline within a retrieval-augmented generation (RAG) module for medical evidence support, addressing the challenge of knowledge currency. Extensive experiments conducted on three EHR datasets demonstrate \modelname{}'s superior performance in clinical mortality outcome and readmission prediction tasks, underscoring its potential to revolutionize clinical decision support systems and advance personalized precision medicine. All code, case studies and a questionnaire are available at the project website: \textcolor{purple}{\url{https://colacare.netlify.app}}.
\end{abstract}

\maketitle
\section{Introduction}
\begin{figure}[!ht]
  \centering
  \includegraphics[width=0.95\linewidth]{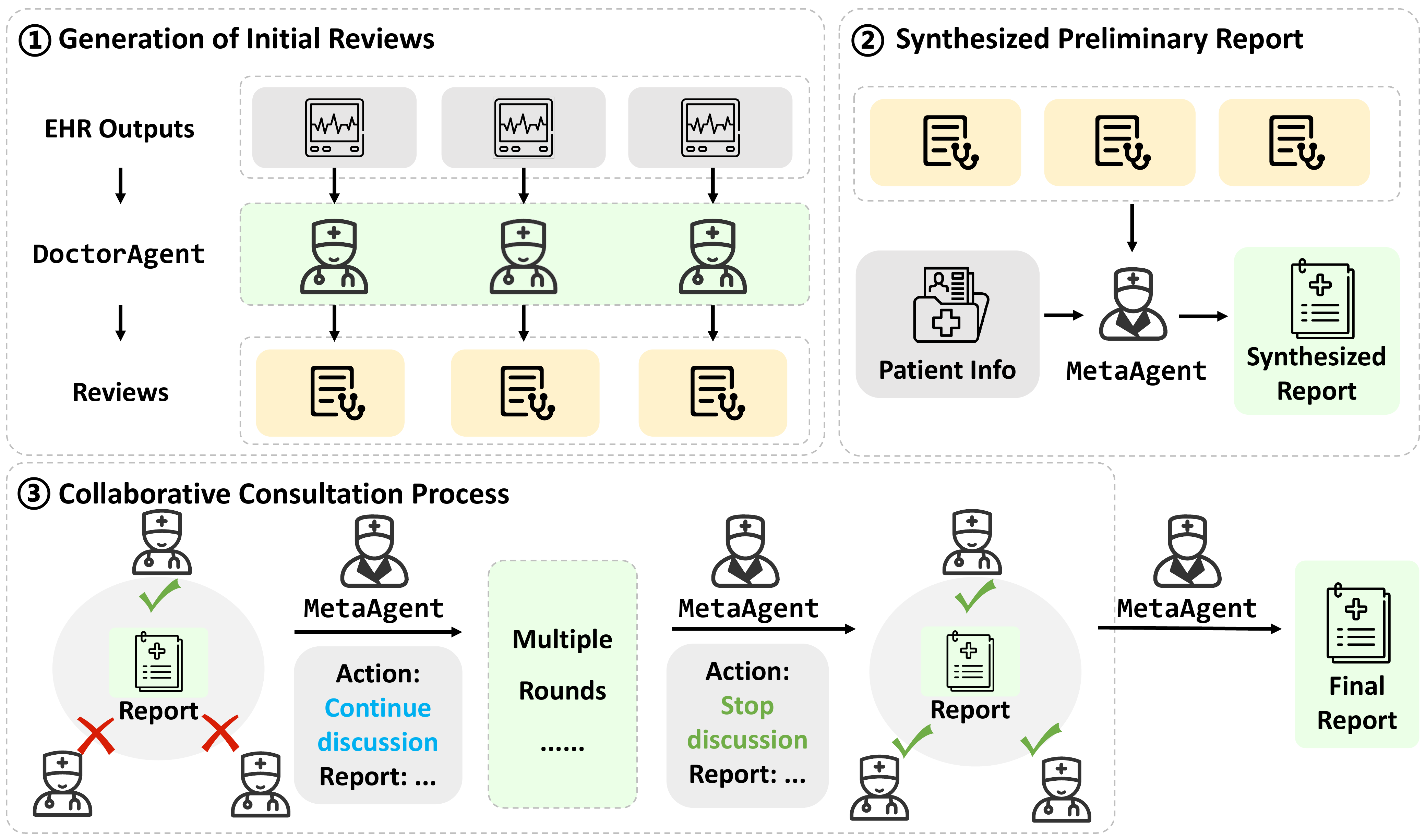}
  \vspace{-3mm}
  \caption{\textit{A three-stage workflow of \modelname{}'s multi-agent collaborative consultation.} (1) multiple \texttt{DoctorAgent}s generate initial reviews based on EHR outputs; (2) a \texttt{MetaAgent} synthesizes these reviews into a preliminary report; and (3) an iterative consultation process where \texttt{DoctorAgent}s and the \texttt{MetaAgent} collaborate through multiple rounds until reaching consensus, resulting in a final report.}
  \label{fig:consultation}
\end{figure}

The Web has become an indispensable platform for facilitating the integration and analysis of Electronic Health Records (EHR) data, playing a pivotal role in prognosis prediction and clinical decision-making. By leveraging web-based technologies, EHR modeling not only enhances data accessibility but also fosters the development of data-driven healthcare advancements~\cite{gao2024comprehensive}. In recent years, deep learning has achieved remarkable success in EHR modeling, particularly through structured data repositories and interoperable platforms that are part of the broader web ecosystem~\cite{ma2020concare, ma2020adacare, zhang2021grasp}. However, these efforts have primarily relied on purely data-driven, end-to-end methods that operate independently of external, web-based semantic knowledge. Consequently, these models often fail to fully capture the clinical significance of the recorded features, treating them as mere variables without a comprehensive semantic context~\cite{zhu2024emerge}. This limitation hinders the real-world applicability of such ``black box'' methods, as they lack interpretability and do not leverage the rich, interlinked knowledge available on the Web, making it difficult for human physicians to trust them in real-world clinical practice.

Most interpretability models primarily rely on traditional feature importance analysis techniques, such as attention mechanisms~\cite{ma2023aicare,zhu2024prism}, SHAP (SHapley Additive exPlanations)~\cite{sundararajan2020many}, and activation-level visualizations~\cite{woo2018cbam}. While these techniques provide a basic level of interpretability, they still fall short of aligning models with real-world knowledge. Several existing methods attempt to incorporate external knowledge to improve interpretability. For instance, some approaches embed knowledge from external knowledge graphs~\cite{choi2017gram, ma2018kame}, or construct knowledge graphs from patient sequential visit data~\cite{jiang2023graphcare}. However, deploying these methods in practical settings remains challenging due to their dependence on manually crafted knowledge representations and the slow pace of knowledge updates, which often fail to keep pace with the latest medical research or updated guidelines. Moreover, physicians increasingly require deeper insights into the reasoning behind deep learning models for interpretability purposes.

Given the impressive capabilities of Large Language Models (LLMs) in handling general tasks~\cite{openai2023gpt4}, including medical applications~\cite{singhal2023llm4clinical} such as medical question-answering (Q\&A) tasks that reason over unstructured clinical notes~\cite{tang2024medagents,hong2024argmed,chen2024reconcile}, we are motivated to explore their potential for enhancing structured EHR modeling, which remains less explored. Specifically, we aim to meet the requirements of knowledge fusion and interpretable reasoning that address the limitations of conventional methods. We have further identified and summarized the key limitations of existing LLM-driven works in structured EHR modeling as follows:
\begin{enumerate}
    \item \textit{Structured EHR comprehension}: Although some works have demonstrated LLMs' reasoning ability in structured EHR analysis under few-shot settings~\cite{han2024evaluatinggpt4}, there remains a notable performance gap compared to conventional methods~\cite{zhu2024benchmarkingllm}. While LLMs excel at mining natural language contexts, their ability to analyze or make predictions based on structured EHR data is still limited~\cite{zhu2024benchmarkingllm}.
    \item \textit{Lack of medical reference}: While recent research has expanded beyond single-model approaches by introducing multiple LLMs to address complex problems~\cite{tang2024medagents,hong2024argmed,chen2024reconcile}, most previous works lack trustworthy medical guidelines for clinical decision-making. These continuously updated guidelines are crucial when the inherent knowledge within LLMs is insufficient for diagnosis.
\end{enumerate}

To address these challenges, we propose \modelname{}, a Retrieval-Augmented Generation (RAG)-enhanced framework inspired by the Multidisciplinary Team (MDT) approach employed by physicians in clinical settings~\cite{pillay2016impact}. \modelname{} integrates domain-specific EHR modeling methods (expert models) into LLM-driven agents that fulfill two roles: doctor agents (referred to as \texttt{DoctorAgent}) and a meta doctor agent (referred to as \texttt{MetaAgent}). The \texttt{MetaAgent} orchestrates MDT discussions, facilitating consultations and evidence-based debates among \texttt{DoctorAgent}s. Overall, \modelname{} simulates the collaborative decision-making process among physicians with diverse expertise by leveraging LLMs' reasoning and role-playing capabilities, alongside the strengths of expert models in processing and predicting from structured EHR data.

\modelname{} has the potential to revolutionize clinical decision-making and advance the field of personalized precision medicine. Our primary contributions are summarized as follows:
\begin{itemize}
    \item \textbf{Insightfully}, we introduce the MDT approach to EHR modeling and incorporate external knowledge via RAG, enabling a prediction process that is both driven by EHR data and enriched with up-to-date external knowledge, complemented by self-examination.
    \item \textbf{Technically}, we develop a collaborative mechanism where clinical decision evidence from multiple \texttt{DoctorAgent}s and the \texttt{MetaAgent} is synthesized. These agents discuss patient status and reach a consensus in a report, thereby enhancing transparency and providing human-understandable evidence to support physicians in their diagnostic reasoning.
    \item \textbf{Experimentally}, we conduct extensive experiments on three EHR datasets, demonstrating the superior performance of \modelname{} in predicting clinical outcomes such as mortality and readmission, with relative improvements in AUPRC of 0.86\%, 2.50\%, 2.00\%, and 4.49\% across four tasks. Case studies highlight the reasonableness and interpretability of the reports generated by \modelname{}, offering healthcare professionals detailed and understandable insights into each prediction.
\end{itemize}

\section{Related Work}

\subsection{LLMs in Medical Tasks}

Large Language Models (LLMs) have demonstrated significant success in the medical domain, particularly in medical question-answering (Q\&A)~\cite{chen2023meditron,xie2024meLlama,feng2024evaluation} and medical evidence summarization tasks~\cite{tang2023evaluating,van2024adapted}.
Notably, the advanced LLM GPT-4~\cite{openai2023gpt4} has outperformed medical students on standard medical board exams~\cite{katz2024gpt}. While these achievements primarily involve textual clinical notes, recent research has begun exploring LLMs' capabilities in handling structured Electronic Health Records (EHR) data~\cite{wu2024ehrflow}. Approaches include prompting LLMs directly~\cite{han2024evaluatinggpt4,zhu2024benchmarkingllm} or ensembling machine learning models' outputs~\cite{hu2024powercombiningdataknowledge,llmforadmission}. These studies reveal that GPT-4 shows potential for zero-shot prediction on structured EHR data, although a significant performance gap remains compared to conventional deep learning methods trained on full datasets~\cite{zhu2024benchmarkingllm}. Additionally, LLM's direct outputs face challenges with hallucination, where generated content may not strictly adhere to instructions, resulting in unexpected outputs beyond numerical values~\cite{xue2023promptcast}.

\subsection{LLM-Driven Multi-Agent Collaboration in Medical Field}

The development of LLM-driven agent systems, where multiple agents with distinct roles collaborate and utilize external tools~\cite{firat2023if, wu2023autogen}, has garnered increasing attention in medical domains. 
AI-SCE~\cite{mehandru2024aisce} introduces a novel evaluation framework that enables the creation of simulated clinical environments for assessing clinical LLM agents. Notable implementations of this approach include AI Hospital~\cite{fan2024aihospital} and Agent Hospital~\cite{li2024agenthospital}, which construct virtual hospital environments where multiple agents collaborate to emulate real clinical settings. These platforms facilitate comprehensive evaluation of clinical LLMs' capabilities in realistic healthcare scenarios.
Recent studies have also explored adversarial collaboration, incorporating debates and negotiations among multiple agents. MedAgents~\cite{tang2024medagents} proposes a medical collaboration framework where doctor agents vote on diagnoses, while ArgMed-Agents~\cite{hong2024argmed} constructs conflict relationship graphs and employs formal deduction to generate coherent Q\&A conclusions. Further advancements include ReConcile~\cite{chen2024reconcile}, which employs a confidence-weighted voting mechanism for better consensus.

In summary, most prior work focuses on medical Q\&A tasks, with LLMs generating text-based diagnoses. However, real-world scenarios often require specific numerical outputs, like disease mortality risk predictions. While approaches like Multi-Agent Debate (MAD)~\cite{guo2024survey,du2023factuality,liang2023mad} have been proposed at a general theoretical level, they are not specifically designed for clinical applications. Moreover, current implementations often use multiple instances of the same LLM, leading to homogeneous reasoning. This falls short of clinical requirements for diverse diagnostic reasoning. Therefore, there is a need for frameworks that better integrate structured EHR and address complex, quantitative medical tasks.

\section{Problem Definition}

\subsection{EHR Datasets Formulation}

The EHR datasets are structured as multivariate time-series data with multiple features, denoted as $\bm{X} = [\bm{x_1}, \bm{x_2}, \cdots, \bm{x_T}]^{\top} \in \mathbb{R}^{T \times F}$, encapsulating information across $T$ visits and $F$ features, which includes static features (e.g., sex and age) and dynamic features (e.g., laboratory tests and vital signs).

\subsection{Predictive Objective Formulation}

The prediction task is defined as a binary classification problem aimed at predicting patients' mortality outcomes or 30-day readmission. Our goal is to extract knowledge from EHR data, supplemented by auxiliary external medical knowledge (e.g., medical guidelines), to enhance predictive modeling of EHRs. Thus, the predictive objective is formulated as:
\begin{equation}
    \hat{y} = \texttt{Framework}(\bm{x}_{EHR}, \textit{MedicalKnowledge})
\end{equation}
where \(\hat{y}\) represents the predicted outcome.

For the mortality prediction task, the outcome \(\hat{y}\) is a binary variable where 0 indicates the patient is alive, and 1 indicates the patient is deceased. In the case of the readmission prediction task, the model predicts whether the patient will be readmitted within 30 days of discharge, with 0 representing no readmission and 1 indicating readmission within the specified time frame.

\section{Methodology}

\begin{figure*}[!ht]
  \centering
  \includegraphics[width=0.8\linewidth]{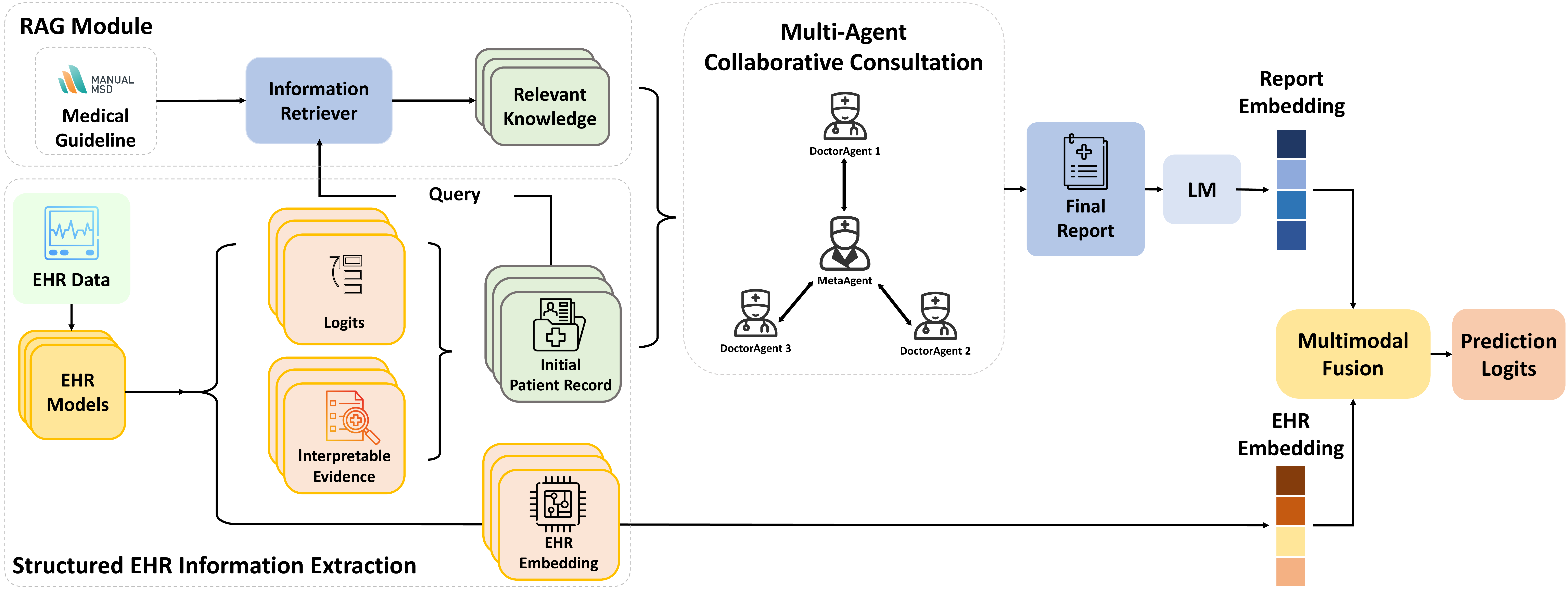}
  \caption{\textit{Overall architecture of our proposed \modelname{} framework.} It consists of three main components: (1) a structured EHR information extraction module that processes patient EHR data and generates embeddings through EHR models, (2) a RAG module that processes medical guidelines and retrieves relevant knowledge, and (3) a multi-agent collaborative consultation module where multiple \texttt{DoctorAgent}s interact with a \texttt{MetaAgent} to produce a final report. The framework concludes with multimodal fusion of report and EHR embeddings to generate prediction logits.}
  \vspace{-3mm}
  \label{fig:overall_pipeline}
\end{figure*}

\cref{fig:consultation} and \cref{fig:overall_pipeline} demonstrates our proposed LLM-based multi-agent collaboration \modelname{} framework, where the key module: Multi-Agent Collaborative Consultation Module is detailedly illustrated in \cref{fig:consultation}. It gathers a group of doctor agents and a meta doctor agent for medical discussion about a certain patient's condition. Doctor agents reach an agreement after multiple rounds of debate and the meta doctor agent proposes a report finally.

\subsection{Structured EHR Information Extraction}
\label{sec:EHR_Model}

Given the EHR data of a patient, $\bm{x}_{EHR}$, we utilize the EHR model, denoted as $\text{Model}$, to encode this temporally linked information:
\begin{equation}
\label{eq:model}
    \bm{h}_{EHR} = \text{Model} (\bm{x}_{EHR})
\end{equation}

Then we use a MLP layer to predict the output logit $\bm{z}$ and the SHAP strategy to obtain feature importance weights $\bm{\alpha}$:
\begin{equation}
\label{eq:ehr_model}
\begin{aligned}
    \bm{z} &= \text{MLP}(\bm{h}_{EHR}) \\
    \bm{\alpha} &= \text{SHAP}(\text{Model}, \bm{x}_{EHR})
\end{aligned}
\end{equation}

The prediction logit $\bm{z}$ and weights $\bm{\alpha}$
by deep learning models are used to support the LLM-based collaborative consultation.

\subsection{Multi-Agent Collaborative Consultation}

As shown in \cref{fig:consultation}, in the multi-agent collaborative consultation, we define two distinct roles: the doctor agent 
(\texttt{DoctorAgent}) and the meta doctor agent (\texttt{MetaAgent}). Each \texttt{DoctorAgent} is linked to a domain-specific expert model. 

The collaboration begins with each \texttt{DoctorAgent} providing an initial review of a patient's condition. Subsequently, the \texttt{MetaAgent} synthesizes these reviews to generate a comprehensive report and orchestrates the collaborative consultation process. During this iterative process, all \texttt{DoctorAgent}s express their opinions on the current report. The \texttt{MetaAgent} then considers the feedback, accordingly revises the report, and determines whether more rounds of consultation are necessary. \cref{fig:consultation} demonstrates the overview pipeline of the collaborative consultation.

\subsubsection{Generation of Initial Reviews}

We first utilize output logits $\bm{z}$ and feature importance weights $\bm{\alpha}$ of EHR models, as well as basic information of a certain patient (e.g. Sex and Age), to build an initial patient record $\bm{x}_{record}$:
\begin{equation}
    \bm{x}_{record} = \textit{Prompt} ( \bm{z}, \bm{\alpha}, \textit{Info}_{\text{patient}})
\end{equation}

Next, the retriever measures the cosine similarity between embeddings of documents in the corpus and the patient record, and select the top-K relevant documents:
\begin{equation}
\label{eq:retriever}
    \bigcup \textit{Docs} = \textit{Retriever}(\bm{x}_{record})
\end{equation}

Lastly, we instruct the \texttt{DoctorAgent} to generate a review:
\begin{equation}
\label{eq:doctor_review}
    \bm{r}_{doctor} = \texttt{DoctorAgent} (\textit{Prompt}(\bm{x}_{record}, \bigcup \textit{Docs}))
\end{equation}

\subsubsection{Synthesized Preliminary Report}
The \texttt{MetaAgent} is instructed to leverage the patient's basic information and initial reviews from all \texttt{DoctorAgent}s to generate a synthesized report:

\begin{equation}
\label{eq:collaboration}
    \bm{r}_{meta} = \texttt{MetaAgent} (Prompt(\sum \bm{r}_{doctor}, Info_{patient}))
\end{equation}

For the report, the \texttt{MetaAgent} is instructed to first assess and articulate the patient's mortality risk, categorizing it as either high or low and then select and incorporate pertinent comments and supporting evidence from the \texttt{DoctorAgent}s' reviews.

\subsubsection{Collaborative Consultation Process}
The \texttt{DoctorAgent}s are instructed to evaluate their initial assessments and state their agreement or disagreement with the current report. In cases of disagreement, they are required to provide detailed rationales and support their arguments with relevant documents retrieved from the corpus:
\begin{equation}
\label{eq:co_da}
    \bm{r}_{doctor}^{j} = \texttt{DoctorAgent} (\textit{Prompt}(\bm{r}_{doctor}^{j-1}, 
    \bm{r}_{meta}^{j-1},
    \bigcup{Docs}))
\end{equation}
where $j$ stands for the j-th round in the process.

Subsequently, the \texttt{MetaAgent} aggregates and analyzes the feedback from all \texttt{DoctorAgent}s to determine whether further discussion is necessary:
\begin{equation}
\label{eq:action}
    \textit{Action}^{j} = \texttt{MetaAgent} (\textit{Prompt}(\sum \bm{r}_{doctor}^{j-1}))
\end{equation}

The \texttt{MetaAgent} evaluates the statements of the \texttt{DoctorAgent}s, focusing on their agreement or disagreement with the current report. If unanimous agreement is reached, the \texttt{MetaAgent} concludes that further discussion is unnecessary. However, in cases of disagreement, the \texttt{MetaAgent} conducts a more detailed analysis by carefully examining the evidence presented by the dissenting \texttt{DoctorAgent}s and assessing the validity and relevance of this evidence within the context of the current report. If any opposing statements and their supporting evidence are considered meritorious, the \texttt{MetaAgent} continues the discussion. 

If the action infers a new round, the \texttt{MetaAgent} considers the opinions and relevant documents by all \texttt{DoctorAgent}s and refine the report:
\begin{equation}
\label{eq:co_ma}
    \bm{r}_{meta}^{j} = \texttt{MetaAgent} (\textit{Prompt}(\bm{r}_{meta}^{j-1}, \sum \bm{r}_{doctor}^{j-1}))
\end{equation}

Once again, \texttt{DoctorAgent}s express their statements towards the new report and try to convince the \texttt{MetaAgent}. The \texttt{MetaAgent} considers their statements and further revises the report, until all \texttt{DoctorAgent}s reach an agreement or the \texttt{MetaAgent} calls it a day.

\subsection{Multimodal Fusion Network}

We begin by leveraging hidden representations of EHR data from previous $N$ EHR-specific expert models, denoted as $\bm{h}_{EHR}^{i}$. Subsequently, we utilize a medical-domain pretrained language model, $\text{LM}$, to encode the final consensus report embedding:
\vspace{-1mm}
\begin{equation}
    \bm{h}_{Report} = \text{LM}(\bm{x}_{Report})
\end{equation}

We then concatenate the representations from both modalities and apply an MLP layer to obtain the final prediction, $\bm{\hat{y}}$:
\begin{equation}
    \bm{\hat{y}} = \text{MLP}\left(\text{Concat}\left[\bm{h}_{EHR}^{1}, \bm{h}_{EHR}^{2}, \ldots, \bm{h}_{EHR}^{N}, \bm{h}_{Report}\right]\right)
\end{equation}

The loss function employed is the Binary Cross-Entropy (BCE) Loss for binary classification:
\begin{equation}
    \mathcal{L}(\bm{\hat{y}}, \bm{y}) = -\frac{1}{N}\sum_{i=1}^{N}(\bm{y}_i \log(\bm{\hat{y}}_i) + (1 - \bm{y}_i) \log(1 - \bm{\hat{y}}_i))
\end{equation}
where $N$ represents the number of patients in a batch, $\bm{\hat{y}} \in [0,1]$ is the predicted probability, and $\bm{y}$ is the ground truth.
\section{Experimental Setups}

\subsection{Experimented Datasets and Utilized Medical Guideline}

We adopt three real-world datasets: (1) \textbf{MIMIC-IV}~\cite{johnson2023mimic} is a comprehensive intensive care unit database containing EHR data including demographics and laboratory test features; (2) \textbf{CDSL}~\cite{cdsl} contains detailed medical information of COVID-19 patients from HM Hospitales; and (3) \textbf{ESRD}~\cite{ma2023aicare} includes longitudinal EHR data from end-stage renal disease (ESRD) patients, with baseline information, visit records and clinical outcomes.

We additionally introduce the Merck Manual of Diagnosis and Therapy (MSD) medical guideline~\cite{porter2011merck}\footnote{Access link: \url{https://www.msdmanuals.com/professional}} to enhance our framework.

We adhere to the established EHR benchmark pipeline~\cite{gao2024comprehensive,ma2023aicare,harutyunyan2019multitask,zhu2024pyehr} for preprocessing time-series data, ensuring consistency and comparability in our data preparation across all datasets. The test set size is limited to approximately 1,000 samples due to the time cost associated with each sample. We suggest a sample of 1,000 patients provides sufficient representation for benchmarking and evaluation, offering diverse, information-rich records that effectively test model generalization in realistic healthcare scenarios~\cite{zhu2024benchmarkingllm,zhu2024prompting}. The statistics of dataset splits and label distributions for the three datasets are presented in \cref{tab:statistics_datasets}.

\begin{table}[!ht]
\footnotesize
\centering
\vspace{-2mm}
\caption{\textit{Statistics of the experimented datasets after preprocessing.} The \textbf{\# Samples} column shows the number of samples and their percentage of the entire dataset, indicating data splits (train, val, test). The \textbf{\# $\text{Label}_{Out.} = 1$} and \textbf{\# $\text{Label}_{Re.} = 1$} columns provides the count and percentage of patients with adverse outcomes within each data split. ``Out.'' denotes ``mortality outcome'', ``Re.'' denotes ``Readmission''.}
\vspace{-2mm}
\label{tab:statistics_datasets}
\begin{tabular}{ccccc}
\toprule
\textbf{Dataset}           & \textbf{Split} & \textbf{\# Samples} & \textbf{\# $\text{Label}_{Out.} = 1$} & \textbf{\# $\text{Label}_{Re.} = 1$} \\ \midrule
\multirow{3}{*}{MIMIC-IV}  & Train          & 17,397 (90.00\%)   & 2,067 (11.88\%)    & 2,685 (15.43\%)    \\
                           & Val            & 966 (5.00\%)      & 118 (12.22\%)     & 157 (16.25\%)     \\
                           & Test           & 968 (5.01\%)      & 115 (11.88\%)     & 153 (15.81\%)     \\ 
\midrule
\multirow{3}{*}{CDSL}      & Train          & 2,127 (49.98\%)    & 270 (12.69\%)     & -     \\
                           & Val            & 1,064 (25.01\%)    & 135 (12.69\%)     & -     \\
                           & Test           & 1,064 (25.01\%)    & 135 (12.69\%)     & -     \\ 
\midrule
\multirow{3}{*}{ESRD}      & Train          & 379 (57.77\%)     & 157 (41.42\%)     & -     \\
                           & Val            & 131 (19.97\%)     & 47 (35.88\%)      & -     \\
                           & Test           & 146 (22.26\%)     & 57 (39.04\%)      & -     \\ 
\bottomrule
\end{tabular}
\vspace{-5mm}
\end{table}

\begin{table*}[!ht]
\footnotesize
\centering
\caption{\textit{Overall performance of in-hospital mortality and 30-day readmission prediction results on MIMIC-IV, CDSL, and ESRD.}}
\vspace{-3mm}
\label{tab:overall_performance}
\resizebox{\textwidth}{!}{\begin{tabular}{l|ccc|ccc|ccc|ccc}
\toprule
\multicolumn{1}{c|}{\multirow{2}{*}{\textbf{Methods}}} & \multicolumn{3}{c|}{\textbf{MIMIC-IV Outcome}} & \multicolumn{3}{c|}{\textbf{MIMIC-IV Readmission}} & \multicolumn{3}{c|}{\textbf{CDSL Outcome}} & \multicolumn{3}{c}{\textbf{ESRD Outcome}} \\
& AUPRC ($\uparrow$) & AUROC ($\uparrow$) & min(+P, Se) ($\uparrow$) 
& AUPRC ($\uparrow$) & AUROC ($\uparrow$) & min(+P, Se) ($\uparrow$) 
& AUPRC ($\uparrow$) & AUROC ($\uparrow$) & min(+P, Se) ($\uparrow$) 
& AUPRC ($\uparrow$) & AUROC ($\uparrow$) & min(+P, Se) ($\uparrow$) \\ 
\midrule
AdaCare & 52.67±4.50 & 87.56±1.37 & 53.62±3.91 & 50.51±4.24 & 78.82±1.94 & 50.01±3.37 & 82.76±3.74 & 95.54±1.02 & 77.20±3.23 & 60.44±6.94 & 69.11±5.53 & 59.23±6.46 \\
ConCare & 49.71±4.83 & 87.21±1.41 & 52.96±3.74 & 46.66±4.05 & 79.06±1.82 & 47.00±3.20 & 80.77±3.59 & 94.47±1.24 & 74.17±2.80 & 56.65±6.91 & 68.84±4.85 & 58.47±5.78 \\
RETAIN & 51.89±4.22 & 87.87±1.27 & 49.71±3.83 & 50.89±3.90 & 80.73±1.82 & 50.42±3.33 & 74.46±4.13 & 93.32±1.30 & 70.09±3.30 & 60.02±6.32 & 70.82±4.32 & 59.26±5.41 \\
\midrule
Ensemble$_\text{Mean}$ & 53.59±4.52 & 88.60±1.26 & 54.08±3.66 & 50.48±4.68 & 86.97±1.45 & 50.36±3.58 & 83.54±3.36 & 95.63±1.04 & 76.24±3.48 & 60.90±7.08 & 73.41±4.80 & 59.98±6.02 \\
Ensemble$_\text{WeightedMean}$ & 53.44±4.51 & 88.58±1.26 & 53.98±3.72 & 50.64±4.69 & 87.05±1.45 & 50.68±3.63 & 84.08±3.34 & 95.88±0.98 & 76.48±3.42 & 61.07±7.11 & 73.49±4.81 & 60.09±5.99 \\
Ensemble$_\text{Temperature}$ & 53.56±4.52 & 88.57±1.26 & 54.07±3.69 & 51.11±4.00 & 80.38±1.80 & \textbf{52.70±3.30} & 83.86±3.32 & 95.72±1.02 & 76.39±3.43 & 61.47±6.41 & 73.36±3.70 & 59.00±5.19 \\
Ensemble$_\text{Deep}$ & 53.12±4.69 & 88.21±1.31 & 53.95±3.57 & 51.54±4.05 & 80.83±1.80 & 52.47±3.27 & 84.58±3.27 & 95.90±0.97 & 77.01±3.38 & 62.61±6.05 & \textbf{73.76±3.76} & 60.47±5.29 \\
MC-Dropout & 52.75±4.93 & 87.96±2.22 & 53.58±3.89 & 50.54±4.66 & 81.57±2.60 & 49.42±3.93 & 83.86±3.63 & 95.04±1.66 & 76.02±3.89 & 61.53±6.90 & 71.61±4.76 & 60.34±5.43 \\
\midrule
LLM$_\text{ZeroShot}$ & 31.01±3.68 & 75.20±2.67 & 34.39±3.71 & 21.63±1.89 & 63.98±2.08 & 26.47±2.71 & 29.71±2.82 & 78.80±2.12 & 33.39±2.63 & 6.71±2.88 & 49.93±12.30 & 7.72±3.28 \\
LLM$_\text{FewShot}$ & 26.28±2.91 & 70.94±2.30 & 40.39±3.74 & 29.09±3.25 & 65.72±1.83 & 34.43±3.62 & 41.58±4.13 & 78.61±1.83 & 42.88±4.53 & 2.78±2.54 & 54.54±38.33 & 2.78±2.54 \\
LLM$_\text{SelfConsistency}$ & 27.39±3.38 & 72.36±2.56 & 30.77±3.55 & 24.73±2.33 & 62.84±2.46 & 27.95±2.79 & 36.24±4.09 & 75.37±2.07 & 32.44±3.96 & 9.09±4.44 & 58.57±10.70 & 11.60±6.06 \\
\midrule
MAD & 30.95±5.59 & 78.43±3.09 & 29.62±4.48 & 24.33±2.66 & 65.09±2.47 & 24.14±2.38 & 42.53±5.38 & 78.55±2.94 & 40.02±5.05 & 5.14±2.63 & 52.47±18.77 & 6.59±3.47 \\
MedAgents & 27.73±3.38 & 74.38±2.19 & 27.80±3.31 & 20.15±2.46 & 58.25±2.66 & 19.39±2.29 & 33.02±3.34 & 81.21±1.75 & 33.48±3.12 & 4.47±2.21 & 53.64±18.31 & 5.26±2.80\\
ReConcile & 27.91±3.50 & 74.51±2.43 & 29.91±3.99 & 24.00±2.86 & 65.39±2.35 & 25.45±3.11 & 35.04±3.17 & 80.39±1.87 & 34.31±2.86 & 17.37±13.54 & 62.20±20.94 & 24.62±16.54 \\
\midrule
\modelname{} & \textbf{54.05±4.88} & \textbf{88.80±1.31} & \textbf{54.44±3.80} & \textbf{51.91±4.99} & \textbf{87.72±1.37} & 50.16±4.17 & \textbf{85.76±2.99} & \textbf{96.23±0.91} & \textbf{78.93±3.21} & \textbf{63.81±6.43} & 71.61±4.77 & \textbf{63.33±5.54} \\ 
\bottomrule
\end{tabular}}
\vspace{-1mm}
\end{table*}

\subsection{Evaluation Metrics}

We employ three widely-used evaluation metrics for binary classification tasks, all of which are interpreted as ``higher is better'':
\begin{itemize}
    \item \textbf{AUROC, AUPRC}: These complementary metrics assess model performance across various classification thresholds. AUROC is valued in clinical settings~\cite{auroc_better}, while AUPRC is particularly useful for imbalanced datasets~\cite{kim2022auprc}. Together, they provide a comprehensive view of the model's discriminative ability.
    \item \textbf{min(+P, Se)}: This metric takes the minimum value between precision (+P) and sensitivity (Se), offering a balanced assessment of model performance by correctly identifying positive cases and minimizing false positives~\cite{ma2022safari}.
\end{itemize}

For readability purposes, in the performance tables below, \textbf{bold} indicates the best results. Values are presented as mean±std, with metrics multiplied by 100.

\subsection{Baseline Models}

\subsubsection{EHR-specific Baselines}

We include the following established EHR-specific deep learning-based models as baselines: AdaCare~\cite{ma2020adacare}, ConCare~\cite{ma2020concare}, and RETAIN~\cite{choi2016retain}. These models employ various architecture, such as attention mechanisms, feature extraction, and recalibration, to address different aspects of EHR data analysis and patient health representation.

Additionally, we implement five ensemble-based approaches: (1) simple averaging of logits from three models; (2) weighted averaging with weights based on model performance; (3) temperature scaling ensemble~\cite{temperature} with trainable parameters; (4) deep ensemble~\cite{deepensemble} which trains each model multiple times with different random initializations and averages their predictions; and (5) MC-dropout~\cite{mcdropout} which enables dropout during inference to generate multiple stochastic predictions for ensemble averaging.

\subsubsection{LLM-based Baselines}

We incorporate baselines utilizing single or multiple LLMs. For those using a single LLM, we employ three different prompt strategies: zero-shot, few-shot, and self-consistency~\cite{wang2022selfconsistency}. Additionally, we include three multi-agent collaboration approaches: multi-agent debate~\cite{liang2023mad}, MedAgents~\cite{tang2024medagents}, and ReConcile~\cite{chen2024reconcile}. All of these LLM-based baselines directly process EHR data as input, relying on large language models to interpret the data and generate final prediction results.

\subsection{Implementation Details}

\subsubsection{Hardware and Software Configuration}

All experiments are conducted on a single Nvidia RTX 3090 GPU with CUDA 12.5. The server's RAM size is 128GB. We implement the model in Python 3.9.19, PyTorch 2.3.1~\cite{paszke2019pytorch}, PyTorch Lightning 2.3.3~\cite{falcon2019lightning}.

\subsubsection{Model Training and Hyperparameters}

AdamW~\cite{loshchilov2017decoupled} optimizer is employed with a batch size of 128 patients. All models are trained for 50 epochs with an early stopping strategy based on AUPRC after 10 epochs without improvement.

For EHR-specific baseline models, the learning rate \{0.01, 0.001, 0.0001\} and hidden dimensions \{64, 128\} are tuned using a grid search strategy on the validation set. For LLM-based baselines, the few-shot approach simulates two sample examples, one positive and one negative, following the approach in previous work~\cite{zhu2024benchmarkingllm}. The self-consistency method selects three possible reasoning paths, followed by a consistency evaluation.

For \modelname{}, the maximum number of rounds for the collaborative consultation process is set to $3$. The fusion network's hidden dimension is $128$ with a learning rate of $0.001$. The $K$ for the retrieval process is set to $3$. Performance is reported as mean±std, calculated by bootstrapping all test set samples 100 times for all three datasets. Experiments are conducted from June 30th, 2024, to October 15th, 2024.

\subsubsection{Utilized (Large) Language Models}

\modelname{} incorporates both Language Models (LMs) and Large Language Models (LLMs) within its framework. For language models, we employ MedCPT~\cite{jin2023medcpt} in the RAG system's biomedical information retrieval process and GatorTron~\cite{yang2022gatortron} model for final clinical note embedding computation. For large language models, we employ the DeepSeek-V2.5~\cite{deepseekai2024deepseekv2} to act as the reasoning engine.

\section{Experimental Results and Analysis}

This section evaluates the \modelname{} framework by addressing the following research questions (\textbf{RQs}):

\begin{enumerate}
    \item \textbf{[RQ1: Overall Performance]} How does \modelname{} compare to other EHR-specific deep learning models and LLM-based frameworks in clinical downstream tasks?
    \item \textbf{[RQ2: Ablation Study]} What is the contribution of each proposed module to the performance?
    \item \textbf{[RQ3: Sensitivity to Number of Agents]} How does the number of agents affect \modelname{}'s performance?
    \item \textbf{[RQ4: Sensitivity to Different LLMs]} How does \modelname{}'s performance vary with different LLMs?
    \item \textbf{[RQ5: Case Study]} Does \modelname{} generate reliable clinical reports and summaries for interpretability?
    \item \textbf{[RQ6: Cost Analysis]} What are the implementation costs of \modelname{} in clinical practice, considering token usage, etc., across datasets and tasks?
    \item \textbf{[RQ7: Human Evaluation]} How consistent are \modelname{}'s decisions and interpretations with experienced clinicians' judgments in real clinical cases?
\end{enumerate}

\subsection{RQ1: Overall Performance}

To address RQ1, we conduct mortality and readmission prediction tasks on the MIMIC-IV dataset and mortality prediction tasks on the CDSL and ESRD datasets. The overall performance of \modelname{} is shown in \cref{tab:overall_performance}. The results demonstrate that \modelname{} consistently outperforms all baseline models in most cases, with notable improvements in the AUPRC metric. Specifically, \modelname{} surpasses each expert model, the ensemble results of expert models (as \modelname{} can be viewed as an LLM-based ensemble method), and LLM-driven approaches that are instructed to directly output prediction results. This superior performance highlights \modelname{}'s potential for practical application in clinical decision-making.

\subsection{RQ2: Ablation Study}

For RQ2, we assess each module's impact on performance (see \cref{tab:ablation_study}) by testing three variants: (1) without the RAG module, where relevant documents are not provided to agents; (2) without the fusion network, which relies on the LLM by prompting it to produce prediction results based on the final report; (3) without integrating expert models, where information from EHR-specific domain models is omitted and the LLM is prompted to generate prediction results based on raw EHR data, replicating the MedAgents approach; and (4) without the MDT process, where relevant documents are integrated into EHR-specific models directly. The results indicate that excluding expert models significantly reduces performance. Additionally, the direct output approach of LLMs proves ineffective, as evidenced by the performance drop when the fusion network is removed and in comparison to other LLM-based baseline methods. Furthermore, \modelname{} performs better compared to the without-MDT baseline and without-RAG baseline, demonstrating that the MDT process in \modelname{}'s framework plays a crucial role and that retrieving relevant documents improves performance.

\begin{table}[h]
\footnotesize
\vspace{-2mm}
\caption{\textit{Ablation study results for each module.}}
\vspace{-2mm}
\label{tab:ablation_study}
\resizebox{\linewidth}{!}{\begin{tabular}{l|ccc|ccc}
\toprule
\multicolumn{1}{c|}{\multirow{2}{*}{\textbf{Methods}}} & \multicolumn{3}{c|}{\textbf{MIMIC-IV Outcome}} & \multicolumn{3}{c}{\textbf{MIMIC-IV Readmission}} \\
& AUPRC ($\uparrow$) & AUROC ($\uparrow$) & min(+P, Se) ($\uparrow$) & AUPRC ($\uparrow$) & AUROC ($\uparrow$) & min(+P, Se) ($\uparrow$) \\
\midrule
w/o RAG & 51.57±4.76 & 88.09±1.30 & 49.25±3.91 & 48.00±4.58 & 86.74±1.63 & \textbf{51.43±3.85} \\
w/o Fusion Network & 42.61±4.02 & 87.53±1.27 & 49.72±3.69 & 49.03±4.38 & 79.85±1.90 & 51.26±3.30 \\
w/o Expert Models & 27.73±3.38 & 74.38±2.19 & 27.80±3.31 & 20.15±2.46 & 58.25±2.66 & 19.39±2.29 \\
w/o MDT & 50.63±5.02 & 86.80±1.36 & 51.92±4.25 & 47.68±4.36 & 79.12±1.82 & 51.39±3.34 \\
\midrule
\modelname{} & \textbf{54.05±4.88} & \textbf{88.80±1.31} & \textbf{54.44±3.80} & \textbf{51.91±4.99} & \textbf{87.72±1.37} & 50.16±4.17 \\
\bottomrule
\end{tabular}}
\vspace{-3mm}
\end{table}

\begin{table}[b]
\footnotesize
\centering
\vspace{-2mm}
\caption{\textit{Performance of different numbers of agents in the in-hospital mortality prediction task on MIMIC-IV datasets.}}
\vspace{-2mm}
\label{tab:number_agents}
\resizebox{\linewidth}{!}{\begin{tabular}{c|ccc|ccc}
\toprule
\multirow{2}{*}{\textbf{\#Agents}} & \multicolumn{3}{c|}{\textbf{Models}} & \multicolumn{3}{c}{\textbf{Metrics}} \\
 & AdaCare & ConCare & RETAIN & AUPRC ($\uparrow$) & AUROC ($\uparrow$) & min(+P, Se) \\
\midrule
\multirow{3}{*}{0} 
    & \checkmark & - & - & 52.67±4.50 & 87.56±1.37 & 53.62±3.91 \\
    & - & \checkmark & - & 49.71±4.83 & 87.21±1.41 & 52.96±3.74 \\
    & - & - & \checkmark & 51.89±4.22 & 87.87±1.27 & 49.71±3.83 \\ \midrule
\multirow{3}{*}{1} 
    & \checkmark & - & - & 52.96±4.54 & 87.59±1.34 & 53.79±3.86 \\
    & - & \checkmark & - & 50.55±4.67 & 87.24±1.42 & 52.78±3.56 \\
    & - & - & \checkmark & 50.44±4.61 & 87.31±1.34 & 49.39±4.02 \\ \midrule
\multirow{3}{*}{2} 
    & \checkmark & \checkmark & - & 52.61±4.85 & 87.96±1.33 & 53.96±3.8 \\
    & \checkmark & - & \checkmark & 51.24±4.60 & 86.58±1.53 & 50.99±3.50 \\
    & - & \checkmark & \checkmark & 52.14±4.74 & 87.91±1.34 & 51.98±3.94 \\ \midrule
3   & \checkmark & \checkmark & \checkmark & \textbf{54.05±4.88} & \textbf{88.80±1.31} & \textbf{54.44±3.80} \\
\bottomrule
\end{tabular}}
\end{table}

\subsection{RQ3: Sensitivity to Number of Agents}

For RQ3, we explore the impact of the number of agents participating in the collaborative consultation, as shown in \cref{tab:number_agents}. Our findings indicate that having one or two \texttt{DoctorAgent}s has little to no impact or slightly decreases performance. This is because a small number of \texttt{DoctorAgent}s can easily reach a consensus, even if their final opinion lacks robustness. As the number of \texttt{DoctorAgent}s increases, performance improves due to the incorporation of a wider range of perspectives and medical evidence, resulting in more comprehensive and reliable reports.

\subsection{RQ4: Sensitivity to Different LLMs}

To address RQ4, we evaluate the performance of \modelname{} when utilizing different LLMs as the reasoning engine. Specifically, we compare \modelname{} instantiated with DeepSeek-V2.5, GPT-4o-Mini, GPT-4o, Qwen-Turbo, Doubao-Pro, Llama-3.1-400B and Claude-3.5-Sonnet-1022. The performance presented in \cref{tab:different_llms} shows that all these LLMs can reason with EHR data, with DeepSeek-V2.5 and GPT-4o-Mini performing slightly better.

\begin{table}[h]
\footnotesize
\vspace{-2mm}
\caption{\textit{Performance of different LLMs of in-hospital mortality and 30-day readmission prediction results on MIMIC-IV.}}
\vspace{-2mm}
\label{tab:different_llms}
\resizebox{\linewidth}{!}{\begin{tabular}{c|ccc|ccc}
\toprule
\multirow{2}{*}{\textbf{Methods}} 
& \multicolumn{3}{c|}{\textbf{MIMIC-IV Outcome}} & \multicolumn{3}{c}{\textbf{MIMIC-IV Readmission}} \\
& AUPRC ($\uparrow$) & AUROC ($\uparrow$) & min(+P, Se) ($\uparrow$) & AUPRC ($\uparrow$) & AUROC ($\uparrow$) & min(+P, Se) ($\uparrow$) \\
\midrule
GPT-4o-Mini & \textbf{55.27±4.92} & 88.66±1.38 & \textbf{55.74±3.75} & 51.01±5.05 & 85.00±1.88 & 51.09±4.09 \\
GPT-4o      & 54.13±4.84 & 88.59±1.36 & 54.38±3.96 & 50.93±4.81 & 86.89±1.48 & 49.44±3.63 \\
Qwen-Turbo  & 52.88±4.76 & 88.57±1.28 & 54.68±3.70 & 49.95±4.13 & 80.10±1.81 & 50.38±3.29 \\
Doubao-Pro  & 52.14±4.65 & 88.04±1.34 & 53.37±3.69 & 49.70±4.04 & 79.92±1.80 & 50.00±3.33 \\
Llama-3 & 54.23±4.81 & 88.20±1.32 & 52.10±3.74 & 51.16±4.08 & 85.35±1.77 & \textbf{51.96±3.05} \\
Claude-3.5-sonnet & 52.58±4.90 & 88.07±1.37 & 53.92±3.63 & 50.77±4.45 & 81.53±1.74 & 50.50±3.45 \\
\midrule
\modelname{} (DeepSeek-V2.5) & 54.05±4.88 & \textbf{88.80±1.31} & 54.44±3.80 & \textbf{51.91±4.99} & \textbf{87.72±1.37} & 50.16±4.17 \\
\bottomrule
\end{tabular}}
\vspace{-4mm}
\end{table}

\subsection{RQ5: Case Study}

\begin{figure*}[!ht]
    \centering
    \includegraphics[width=0.8\linewidth]{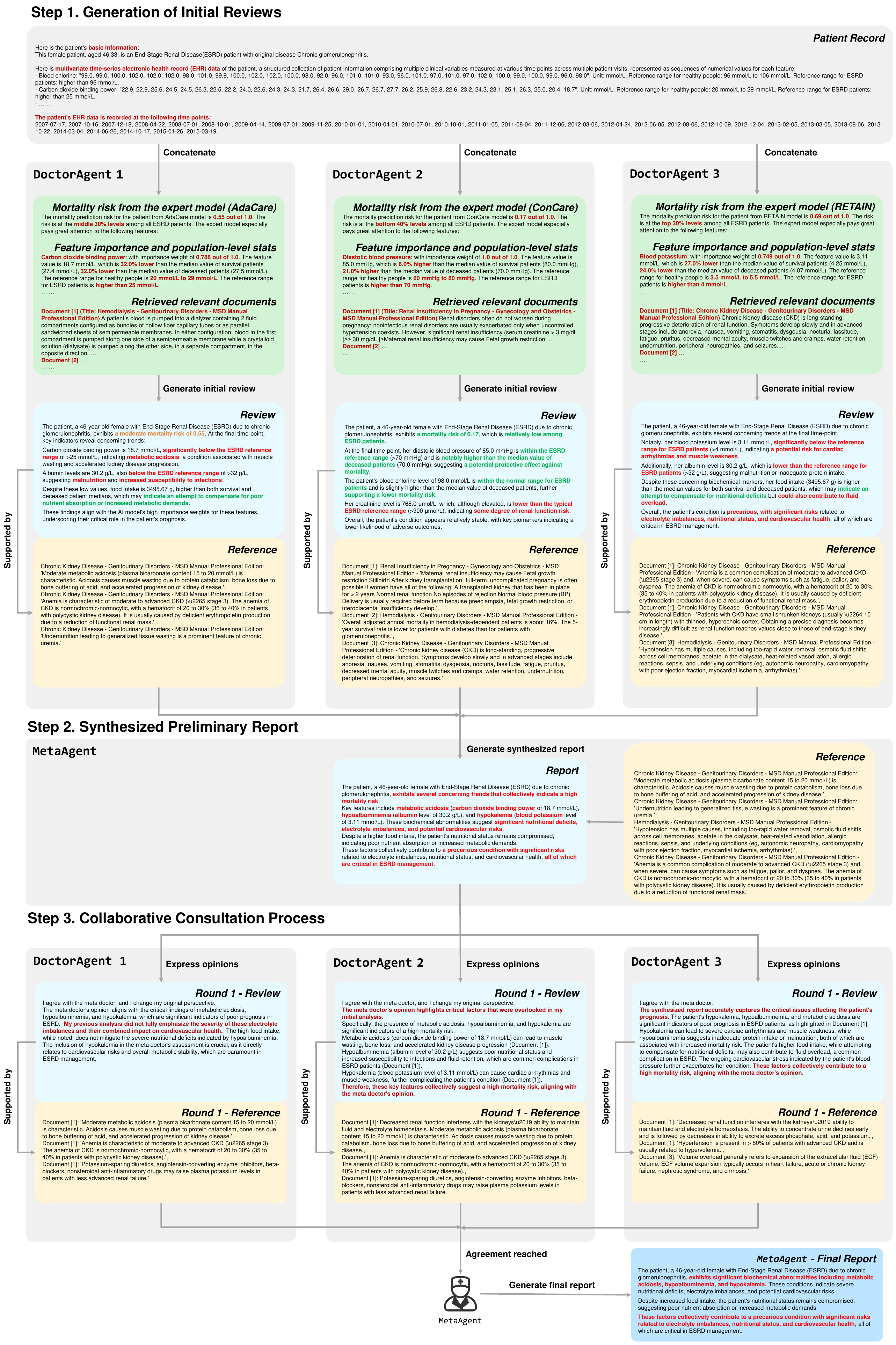}
    \caption{\textit{The case of a patient from ESRD dataset, who deceased within one year after the last follow-up visit.} Important indicators are shown in dark red color. Diseases are shown in red color. Healthy indicators are shown in green color.}
    \label{fig:case_study}
\end{figure*}

To evaluate whether \modelname{} provides reliable clinical reports and summaries for interpretable analysis, we present a case study based on a patient from the ESRD dataset's mortality prediction task. \cref{fig:case_study} illustrates each step in the \modelname{} pipeline during the collaborative consultation process.

In Step 1, each \texttt{DoctorAgent} is provided with the patient's records, including basic information and multivariate time-series EHR data, results from expert models such as mortality risk prediction logits, feature importance, and population-level statistics, as well as relevant retrieved documents. The \texttt{DoctorAgent}s are then prompted to generate an initial review for the patient. \texttt{DoctorAgent} 1 assesses a moderate risk by focusing on the patient's carbon dioxide binding power and albumin levels. \texttt{DoctorAgent} 2 determines a low risk by concentrating on diastolic blood pressure and blood chlorine levels, both within normal ranges for patients with end-stage renal disease (ESRD). In contrast, \texttt{DoctorAgent} 3 identifies significantly low blood potassium levels, indicating a precarious condition. All \texttt{DoctorAgent}s cite evidence from authoritative documents. In Step 2, the \texttt{MetaAgent} synthesizes a report based on the reviews from the three \texttt{DoctorAgent}s, primarily integrating the analyses of \texttt{DoctorAgent} 1 and \texttt{DoctorAgent} 3 to conclude a high mortality risk and provide a comprehensive analysis. The \texttt{MetaAgent} highlights key factors such as carbon dioxide binding power, blood potassium, and albumin levels, all of which are abnormal and pose serious risks. Subsequently, in Step 3, each \texttt{DoctorAgent} reviews, votes on, and comments on the \texttt{MetaAgent}'s synthesized report. \texttt{DoctorAgent} 1 and \texttt{DoctorAgent} 2 concur with the \texttt{MetaAgent}'s perspective, revising their initial assessments and addressing previously overlooked critical factors. Upon reaching consensus among all \texttt{DoctorAgent}s, the \texttt{MetaAgent} performs a final summary and delivers the ultimate report.

\subsection{RQ6: Cost Analysis}

To evaluate the practical viability of the \modelname{} framework in real-world clinical settings, we conduct an analysis of its computational costs, as illustrated in \cref{tab:cost_stats}. With DeepSeek-V2.5, \modelname{} costs approximately \$0.013 USD per patient, demonstrating its cost-effectiveness and practical viability for integration into clinical workflows.

\begin{table}[h]
\footnotesize
\centering
\vspace{-1mm}
\caption{\textit{Average time and token cost of \modelname{} in in-hospital mortality and 30-day readmission prediction tasks on MIMIC-IV, CDSL, and ESRD datasets.} }
\vspace{-2mm}
\label{tab:cost_stats}
\resizebox{\linewidth}{!}{\begin{tabular}{c|c|c|c|c}
\toprule
\textbf{Cost} & \textbf{MIMIC-IV Outcome} & \textbf{MIMIC-IV Readmission} & \textbf{CDSL Outcome} & \textbf{ESRD Outcome} \\
\midrule
Prompt Tokens & 59,113.06 & 62,759.91 & 124,074.52 & 72,151.11 \\
Output Tokens & 4,435.91 & 4,570.47 & 5,456.79 & 6,156.30 \\
\# API Requests & 8.56 & 8.37 & 10.39 & 11.86 \\
\bottomrule
\end{tabular}}
\vspace{-4mm}
\end{table}

\subsection{RQ7: Human Evaluation}

We conduct a systematic questionnaire-based human evaluation study using peritoneal dialysis patient data. We engage 12 experienced medical practitioners (5-15 years of experience) from nephrology departments across 5 hospitals. Practitioners rate agreement levels (1-5) for each patient case. \modelname{} achieves a mean score of 4.4 vs. ConCare's 3.2, demonstrating strong alignment with expert clinical judgment. Blinded evaluations confirm that \modelname{}'s interpretability closely matches human expert practice patterns.

\section{Conclusion}

This paper presents \modelname{}, a framework that enhances EHR modeling through LLM-driven multi-agent collaboration. \modelname{} combines a knowledge-infused Multidisciplinary Team ensemble within an LLM-driven multi-agent system, supported by RAG for accessing up-to-date medical knowledge. This design mirrors the collaborative and iterative nature of real-world clinical decision-making processes, delivering human-interpretable, personalized predictions with patient-specific evidence while identifying and rectifying potential errors. Experiments on three real-world EHR datasets demonstrate \modelname{}'s better performance in mortality and readmission predictions. By simulating real clinical team dynamics, \modelname{} advances personalized precision medicine and has the potential to transform clinical decision support systems and healthcare delivery.

\section*{Ethical Use of Data}

All EHR data used in this study are de-identified EHR datasets. We adhere to the data usage policy. The MIMIC-IV dataset is processed using secure Azure OpenAI API and human review of the data has been waived.

\begin{acks}
This work was supported by the National Natural Science Foundation of China (62402017, U23A20468), Beijing Natural Science Foundation (L244063), Xuzhou Scientific Technological Projects (KC23143), Peking University Medicine plus X Pilot Program-Key Technologies R\&D Project (2024YXXLHGG007). Junyi Gao acknowledges the receipt of studentship awards from the Health Data Research UK-The Alan Turing Institute Wellcome PhD Programme in Health Data Science (Grant Ref: 218529/Z/19/Z).
\end{acks}

\clearpage
\newpage
\bibliographystyle{ACM-Reference-Format}
\bibliography{ref}

\appendix
\section{Limitations and Future Work}

While \modelname{} demonstrates promising results in enhancing EHR modeling, several limitations and areas for future work remain:

\begin{itemize}
    \item \textbf{Generalizability:} Our current study focuses on mortality and readmission prediction tasks across three datasets and employs large language models, including DeepSeek-V2.5 and GPT-4o. Further research is needed to evaluate \modelname{}'s performance on a broader range of clinical prediction tasks and diverse EHR datasets. Additionally, incorporating more LLMs, both closed-source and open-source, such as Claude, Gemini, LLaMA, and Qwen, could further validate the framework's generalizability to meet users' flexible needs.
    \item \textbf{Human Evaluation:} Although \modelname{} generates interpretable reports that are examined by expert doctors in our case studies, more extensive human evaluation involving a larger cohort of clinicians is necessary.
    \item \textbf{Continuous Learning:} The current parameterization of EHR-specific models stays fixed after training on full datasets. Developing mechanisms that enable \modelname{} to continuously learn and update its parameters based on feedback from LLM agents and real-world clinical data remains an area to explore to enhance its adaptability and performance over time in dynamic healthcare environments.
\end{itemize}

Our future work aims to address these limitations to further improve the effectiveness and applicability of \modelname{} in real-world clinical settings. By leveraging a broader set of LLMs and introducing mechanisms for confidence estimation and continuous learning, we aim to create a more robust, interpretable, and adaptive tool for EHR modeling and clinical decision-making.

\section{Supplemental Experiments}

We employed M3Care, AICare, and PRISM as ColaCare’s expert models (and compared them individually). Results validate ColaCare’s effectiveness in enhancing performance through DoctorAgents, MetaAgent collaboration, debate, and LLM reasoning. Furthermore, we compared the latest EMERGE model that incorporates external knowledge graphs and uses RAG, demonstrating that ColaCare performs better.

\begin{itemize}
    \item M3Care~\cite{zhang2022m3care}: Learning with Missing Modalities in Multimodal Healthcare Data
    \item AICare~\cite{ma2023aicare}: Mortality prediction with adaptive feature importance recalibration for peritoneal dialysis patients
    \item PRISM~\cite{zhu2024prism}: Mitigating EHR Data Sparsity via Learning from Missing Feature Calibrated Prototype Patient Representations
    \item EMERGE~\cite{zhu2024emerge}: Enhancing Multimodal Electronic Health Records Predictive Modeling with Retrieval-Augmented Generation
\end{itemize}

\begin{table*}[!ht]
\footnotesize
\centering
\caption{\textit{Overall performance of in-hospital mortality and 30-day readmission prediction results on MIMIC-IV, CDSL, and ESRD datasets.} }
\label{tab:sup_performance}
\resizebox{\textwidth}{!}{
\begin{tabular}{c|ccc|ccc|ccc|ccc}
\toprule
\multicolumn{1}{c|}{\multirow{2}{*}{\textbf{Methods}}} & \multicolumn{3}{c|}{\textbf{MIMIC-IV Outcome}} & \multicolumn{3}{c|}{\textbf{MIMIC-IV Readmission}} & \multicolumn{3}{c|}{\textbf{CDSL Outcome}} & \multicolumn{3}{c}{\textbf{ESRD Outcome}} \\
\multicolumn{1}{c|}{} & AUPRC ($\uparrow$) & AUROC ($\uparrow$) & min(+P, Se) ($\uparrow$) & AUPRC ($\uparrow$) & AUROC ($\uparrow$) & min(+P, Se) ($\uparrow$) & AUPRC ($\uparrow$) & AUROC ($\uparrow$) & min(+P, Se) ($\uparrow$) & AUPRC ($\uparrow$) & AUROC ($\uparrow$) & min(+P, Se) ($\uparrow$) \\
\toprule
M3Care & 52.77±4.57 & 87.45±1.28 & 53.26±3.79 & 50.62±4.30 & 80.14±1.84 & 51.59±3.46 & 76.17±3.96 & 93.91±1.19 & 70.69±3.64 & 56.52±6.86 & 71.97±4.12 & 54.81±5.36 \\
AICare & 50.20±4.58 & 87.66±1.41 & 48.77±3.74 & 49.51±4.13 & 80.35±1.84 & 49.87±3.39 & 83.69±3.08 & 94.19±0.98 & 76.78±2.97 & 62.37±6.26 & 75.35±3.86 & 59.23±5.42 \\
PRISM & 52.23±4.22 & 87.68±1.23 & 53.01±3.59 & 50.95±4.05 & 80.38±1.93 & 52.01±3.41 & 84.28±3.14 & 95.16±0.87 & 78.05±2.86 & 62.13±5.84 & 73.11±3.87 & 54.16±4.93 \\
\midrule
Ensemble$_\text{Mean}$ & 54.54±4.45 & 88.80±1.25 & 54.19±3.77 & 51.13±4.18 & 80.90±1.84 & 50.75±3.43 & 86.04±3.19 & 96.71±0.83 & 79.39±2.91 & 63.40±6.20 & 76.90±3.62 & 63.48±4.70 \\
Ensemble$_\text{WeightedMean}$ & 54.61±4.45 & 88.82±1.24 & 54.22±3.81 & 51.12±4.18 & 80.89±1.84 & 50.79±3.43 & 86.19±3.17 & 96.73±0.83 & 79.38±2.98 & 63.16±6.18 & \textbf{77.15±3.61} & \textbf{64.05±4.75} \\
Ensemble$_\text{Temperature}$ & 54.78±4.45 & 88.83±1.24 & 54.47±3.85 & 51.22±4.18 & 80.89±1.85 & 51.07±3.37 & 85.65±3.19 & 96.57±0.85 & 78.79±2.82 & 64.74±6.45 & 74.41±3.71 & 60.89±4.71 \\
Ensemble$_\text{Deep}$ & 54.43±4.55 & 88.41±1.27 & 54.22±3.77 & 51.09±4.15 & 80.67±1.85 & 50.76±3.40 & 86.78±2.92 & 96.83±0.80 & 79.12±2.88 & 65.50±6.18 & 73.96±3.61 & 63.49±4.59 \\
MC-Dropout & 54.62±4.49 & 88.10±1.69 & 54.77±4.00 & 51.02±4.65 & 83.75±2.05 & 50.62±3.99 & 86.29±4.25 & 96.58±1.46 & 78.36±3.41 & 65.62±5.74 & 74.29±4.27 & 63.23±5.37 \\
\midrule
EMERGE & 53.08±4.65 & 88.09±1.26 & 53.94±3.73 & \textbf{51.35±3.89} & 82.27±1.88 & 51.44±3.37 & 81.56±4.34 & 95.92±1.53 & 46.03±19.15 & 56.45±6.80 & 75.14±3.47 & 57.26±4.96 \\
\midrule
\modelname & \textbf{55.24±4.60} & \textbf{89.40±1.20} & \textbf{54.99±3.66} & 50.68±4.08 & \textbf{84.22±1.82} & \textbf{52.51±3.42} & \textbf{87.55±3.28} & \textbf{97.62±0.98} & \textbf{80.68±2.99} & \textbf{68.62±6.23} & 74.51±4.08 & 62.78±5.15 \\
\bottomrule
\end{tabular}
}
\end{table*}

\section{Example Prompt Templates}
\subsection{Prompt Template of \texttt{DoctorAgent}’s Initial Review}

\begin{custommdframed}

1. \textbf{Here is the relevant evidence} ($\bigcup \text{Docs}$): \\
Document [0] (Title: Diabetic Nephropathy - Genitourinary Disorders ...\\
Document [1] ...... \\
... \\
2. \textbf{Here is the patient record, including the patient's basic information and analysis results of AI models} ($\bm{x}_{record}$): \\
\textbf{($Info_{patient}$)} This male patient, aged 65.73, is an End-Stage Renal Disease (ESRD) patient with original disease Diabetic Nephropathy, and basic disease Diabetes ...\\
\textbf{(Logit $\bm{z}$)} The mortality prediction risk for the patient from AdaCare model is \textbf{0.01 out of 1.0}.\\
\textbf{(Feature $\bm{\alpha}$)} We pay great attention to these features:
Carbon dioxide binding power: with shap value of 0.165. The feature value ...\\
3. \textbf{You need to analyze the patient's condition based on the above information and generate an analytical review} ($\bm{r}_{meta}$).
\end{custommdframed}
\subsection{Prompt Template of \texttt{MetaAgent}'s Synthesized Report}

\begin{custommdframed}

1. \textbf{First, please read the patient's basic information carefully} ($Info_{patient}$):\\
This male patient, aged 65.73, is an End-Stage Renal Disease (ESRD) patient with original disease Diabetic Nephropathy, and basic disease Diabetes.
 \\
2. \textbf{All doctors made a diagnosis on the patient's condition and gave their reasons as follows} ($\sum \bm{r}_{doctor}$):\\
\textbf{Doctor 0}:
The mortality risk of the patient is: 0.01. 
The patient's basic condition is: a 65-year-old male with End-Stage Renal Disease due to ...
The retrieved evidence: ...\\
\textbf{Doctor 1}: ...
 \\
3. \textbf{You need to consider all doctors' opinions carefully and write a synthesized report} ($\bm{r}_{meta}$).
\end{custommdframed}
\subsection{Prompt Template of \texttt{DoctorAgent}s’ Collaboration}
\begin{custommdframed}

1. \textbf{Here is your initial review, you may adjust it based on the meta doctor's report} ($\bm{r}_{doctor}^{j-1}$) :\\
The patient's basic condition is: a 65-year-old male with End-Stage Renal Disease due to ...  \\
2. \textbf{Here is the synthesized report generated by the meta doctor} ($\bm{r}_{meta}^{j-1}$): \\
(1) In my opinion, the patient has a high risk of mortality. \\
(2) The patient exhibits significant clinical markers of severe renal dysfunction and metabolic imbalances......
 \\
3. \textbf{Here is the relevant document newly retrieved by the retriever module} ($\bigcup{Docs}$): \\
Document [0] (Title: Hemodialysis ...\\
Document [1] ...
\\
4. \textbf{You need to consider the synthesized report carefully and provide your own opinions} ($\bm{r}_{doctor}^{j}$).
\end{custommdframed}

\subsection{Prompt Template of \texttt{MetaAgent}’s Next Action}

\begin{custommdframed}

1. \textbf{Several doctors put forward their own opinions and reasons} ($\sum \bm{r}_{doctor}^{j-1}$): \\
Doctor 0's statement is:
I \textbf{agree with} the meta doctor.
The reason is:
the patient's condition, as described in the report, aligns with the high risk...\\
Doctor 1's statement is: I \textbf{disagree with} the meta doctor. The reason is ... The evidence is ...\\
2. \textbf{Next, you need to judge whether the next round of discussion is needed based on each doctor's statement} ($\textit{Action}^{j}$).  
\end{custommdframed}
\subsection{Prompt Template of \texttt{MetaAgent}’s Revised Report}


\begin{custommdframed}

1. \textbf{In the previous discussion, you gave the report} ($\bm{r}_{meta}^{j-1}$): \\
The patient has a high risk of mortality.
Key concerns include significantly low systolic and diastolic blood pressures...
 \\
2. \textbf{All the doctors offered new perspectives} ($\sum \bm{r}_{doctor}^{j-1}$): \\
Doctor 0's statement is:
the mortality risk of the patient is 0.15. The patient shows a low mortality risk of ... 
The supporting evidence is: 
Document [3] (Title: Anemia of Renal Disease ...
Document ...\\
Doctor 1's statement is:
...... \\
3. \textbf{You need to consider all the doctors' new ideas and modify your original report} ($\bm{r}_{meta}^{j}$). 
\end{custommdframed}

\section{Detailed Dataset Description}

We adopt three real-world datasets: MIMIC-IV, CDSL, and ESRD datasets and additionally introduce the MSD (Merck Manual of Diagnosis and Therapy) medical guideline to enhance our experimental framework.

\begin{itemize}
    \item \textbf{MIMIC-IV}~\cite{johnson2023mimic}: MIMIC-IV is part of the Medical Information Mart for Intensive Care project. It contains comprehensive EHR data for intensive care unit patients, including demographic information, vital signs, laboratory results, procedures, medications, clinical notes, and mortality statistics. For our study, we focus on the clinical notes, demographic data, and laboratory test features.
    \item \textbf{CDSL}~\cite{cdsl}: This dataset is derived from the HM Hospitales EHR system and consists of anonymized records of 4,479 patients admitted with a confirmed or suspected diagnosis of COVID-19. CDSL offers a rich variety of medical features, including comprehensive details on diagnoses, treatments, admissions, ICU stays, diagnostic imaging tests, laboratory results, and patient discharge or death status.
    \item \textbf{ESRD}~\cite{ma2023aicare}: The end-stage renal disease (ESRD) dataset comprises data from 656 patients, including 13,091 visit records collected over a 12-year period, from January 1, 2006, to January 1, 2018. This longitudinal dataset features patients' baseline information, visit records, and clinical outcomes.
\end{itemize}

We incorporate the Merck Manual of Diagnosis and Therapy (MSD) medical guideline~\cite{porter2011merck} into our framework \modelname{}. The MSD guideline is a comprehensive medical reference that provides detailed information on various diseases, corresponding diagnoses, and treatment protocols.

\end{document}